\pdfoutput=1

\documentclass[11pt]{article}

\usepackage{ACL2023}

\usepackage{times}
\usepackage{latexsym}
\usepackage{bm}

\usepackage{amsmath}
\usepackage{times}
\usepackage{latexsym}
\usepackage{graphicx} 
\usepackage{amsmath}
\usepackage{amsfonts}
\usepackage{bm}
\usepackage[linesnumbered,ruled,vlined]{algorithm2e}
\usepackage{comment}
\usepackage{microtype}
\usepackage{url}

\usepackage{hyperref}
\usepackage{arydshln}
\usepackage{multicol,multirow}

\usepackage[T1]{fontenc}

\usepackage[utf8]{inputenc}

\usepackage{microtype}

\usepackage{inconsolata}

\title{Duplex Diffusion Models Improve Speech-to-Speech Translation\thanks{\ Accepted by \href{https://2023.aclweb.org/}{ACL 2023 Findings}}}

\author{Xianchao Wu \\
  NVIDIA \\
  \texttt{\href{xianchaow@nvidia.com}{xianchaow@nvidia.com}, \href{wuxianchao@gmail.com}{wuxianchao@gmail.com}} \\}

\begin{document}
\maketitle
\begin{abstract}
Speech-to-speech translation is a typical sequence-to-sequence learning task that naturally has two directions. How to effectively leverage bidirectional supervision signals to produce high-fidelity audio for both directions? Existing approaches either train two separate models or a multitask-learned model with low efficiency and inferior performance. In this paper, we propose a duplex diffusion model that applies diffusion probabilistic models to both sides of a reversible duplex Conformer, so that either end can simultaneously input and output a distinct language's speech. Our model enables reversible speech translation by simply flipping the input and output ends. Experiments show that our model achieves the first success of reversible speech translation with significant improvements of ASR-BLEU scores compared with a list of state-of-the-art baselines.
\end{abstract}

\section{Introduction}

Direct speech-to-speech translation (S2ST) \cite{s2s_unit_DBLP:journals/corr/abs-2107-05604, unity_s2s_2022}, transforming a source language's speech to the target language's speech, is essential for online international communications and is friendly to numerous languages that do not have their own writing systems or textual vocabularies. 
S2ST circumvents a cascaded architecture \cite{1997_s2st_599557, 2006_s2st_1597243, book_s2st_Wahlster2000} of combining automatic speech recognition (ASR) of the source speech, textual source-to-target machine translation (MT), and target text-to-speech (TTS) synthesis where multiple types of datasets are required, error propagates, latency is high, and unavailable for thousands of (spoken) languages who do not have a writing system.

For S2ST, speech-to-speech parallel data is required, and it is costly to collect a comparable size dataset with textual counterparts. To alleviate the data scarcity problem, self-supervised pre-training and data augmentation techniques were used by \newcite{popuri2022enhanced}, and unsupervised and weakly-supervised speech and text data under Translatotron 2 \cite{Jia2021Translatotron2H} were leveraged by \newcite{Jia2022LeveragingUA}. Techniques such as multi-task learning \cite{Weiss2017SequencetoSequenceMC}, pseudo labeling \cite{Pino2020SelfTrainingFE}, and knowledge distillation \cite{inaguma-etal-2021-source} have also been adapted and achieved promising results.

From S2ST architecture's point of view, \newcite{unity_s2s_2022} describes four categories, (1) Translatotron \cite{Translatotron_Jia2019DirectST} style which includes a speech encoder and a spectrogram decoder, (2) Translatotron2+ \cite{Jia2021Translatotron2H} style which inserts a first-pass text decoder followed by a TTS encoder between the two modules of Translatotron, (3) speech-to-unit translation (S2UT) \cite{s2s_unit_DBLP:journals/corr/abs-2107-05604} that uses discrete clustered units of the target language speech instead of spectrogram, and (4) UnitY \cite{unity_s2s_2022} that inserts a first-pass text decoder followed by a text-to-unit (T2U) encoder between the two modules in S2UT. 

In this paper, following the motivations of textual duplex machine translation \cite{reder_NEURIPS2021_afecc60f}, we leverage S2ST's two directions: effectively utilizing supervision signals from both directions is estimated to both relieve the pain of data scarcity and bring novel architectures of training and inferencing. Existing architectures (e.g., Translatotron1/2, S2UT, and UnitY) either train two separate models or a multitask-learned model with low efficiency and inferior performance. In contrast, we propose a \emph{duplex diffusion model} that applies diffusion probabilistic models to both sides of a \emph{reversible duplex Conformer}, so that either end can simultaneously input and output a distinct language's speech. Our model enables reversible speech translation by simply flipping the input and output ends. Experiments show that our model achieves the first success of reversible speech translation with significant improvements of ASR-BLEU scores compared with a list of strong baselines.

Our contributions are concluded as follows:
\begin{itemize}
    \item a novel \emph{reversible duplex Conformer} that extends the widely used Conformer \cite{conformer_gulati20_interspeech} architecture from ASR to S2ST, with reversible and symmetrical forward/reverse building blocks;
    \item a novel \emph{duplex diffusion model} that jointly train one reversible duplex Conformer in diffusion ways to fit two translation directions;
    \item significantly better or comparable ASR-BLEU scores are achieved by comparing with a list of state-of-the-art baselines including Translatotron, Translatotron2, S2UT, and UnitY.
\end{itemize}


\section{Backgrounds}

\subsection{REDER}

REDER, \textbf{RE}versible \textbf{D}uplex Transform\textbf{ER}, was proposed by \newcite{reder_NEURIPS2021_afecc60f} for reversible textual machine translation through duplex sequence-to-sequence (seq2seq) learning. A neural network with a parameter set $\theta$ is \emph{duplex} when it satisfies the following conditions. 
First, the network has two ends, each end can take one language as its input or output. Second, the network defines a forward mapping function $f_\theta^\rightarrow: \mathcal{X}\mapsto \mathcal{Y}$, and a backward (reverse) mapping function $f_\theta^\leftarrow: \mathcal{Y}\mapsto \mathcal{X}$, that satisfies two reversibilities: $f_\theta^\leftarrow = (f_\theta^\rightarrow)^{-1}$ and $f_\theta^\rightarrow = (f_\theta^\leftarrow)^{-1}$. Third, the network satisfies the cycle consistencies: $\forall \bm{x} \in \mathcal{X}: f_\theta^\leftarrow(f_\theta^\rightarrow(\bm{x})) = \bm{x}$ and $\forall \bm{y} \in \mathcal{Y}: f_\theta^\rightarrow(f_\theta^\leftarrow(\bm{y})) = \bm{y}$. 

However, building duplex seq2seq networks is non-trivial and must satisfy the following constraints, \textbf{reversibility} and \textbf{homogeneity}. First, vanilla encoder-decoder network, such as frequently used Transformer \cite{transformer_NIPS2017_3f5ee243} and its variants, is irreversible. It is not feasible for the output end of the decoder side to take in input signals to exhibit the encoding functionality and vice versa. Second, the natural network architectures of the non-autoregressive encoder and the autoregressive decoder are heterogeneous. Therefore, REDER, leverages reversible Transformer layers \cite{reversible_DBLP:journals/corr/GomezRUG17} and fully non-autoregressive modeling without explicit encoder and decoder division, is designed to solve these two challenges. As reported in \cite{reder_NEURIPS2021_afecc60f}, REDER worked in a duplex way that better exploited the bidirectional supervisions for achieving better downstream reversible machine translation tasks' performance.

The architecture of REDER is a stack of $L$ reversible duplex transformer layers where the 1-st to $L/2$-th layers are mirror of the $(L/2+1)$-th to $L$-th layers to ensure the whole model being symmetric. In particular, each layer contains a multi-head self-attention (MHSA) module and a feed-forward network (FFN) module with a novel reversible design to ensure duplex behavior, where the input and output tensors of such a layer are split into two halves, $\mathbf{H}_{l-1}=[\mathbf{H}_{l-1}^{(1)}; \mathbf{H}_{l-1}^{(2)}]$ and $\mathbf{H}_{l}=[\mathbf{H}_{l}^{(1)}; \mathbf{H}_{l}^{(2)}]$, respectively. Formally, the \emph{regular form} of the $l$-th layer $\mathcal{F}_l$ performs as follow:
\begin{align}
    [\mathbf{H}_{l}^{(1)}; \mathbf{H}_{l}^{(2)}] & =\mathcal{F}_l([\mathbf{H}_{l-1}^{(1)}; \mathbf{H}_{l-1}^{(2)}]), \\
    \mathbf{H}_{l}^{(1)} & = \mathbf{H}_{l-1}^{(1)} + \text{MHSA}(\mathbf{H}_{l-1}^{(2)}), \\
    \mathbf{H}_{l}^{(2)} & = \mathbf{H}_{l-1}^{(2)} + \text{FFN}(\mathbf{H}_{l}^{(1)}).
\end{align}
The \emph{reverse form} $\mathcal{F}_l^{-1}$ can be computed by subtracting the residuals:
\begin{align}
    [\mathbf{H}_{l-1}^{(1)}; \mathbf{H}_{l-1}^{(2)}] & =\mathcal{F}_l^{-1}([\mathbf{H}_{l}^{(1)}; \mathbf{H}_{l}^{(2)}]), \\
    \mathbf{H}_{l-1}^{(2)} & = \mathbf{H}_{l}^{(2)} - \text{FFN}(\mathbf{H}_{l}^{(1)}), \\
    \mathbf{H}_{l-1}^{(1)} & = \mathbf{H}_{l}^{(1)} - \text{MHSA}(\mathbf{H}_{l-1}^{(2)}).
\end{align}


\subsection{DDPM}

\begin{figure}[t]
  \centering
  \includegraphics[width=7.5cm]{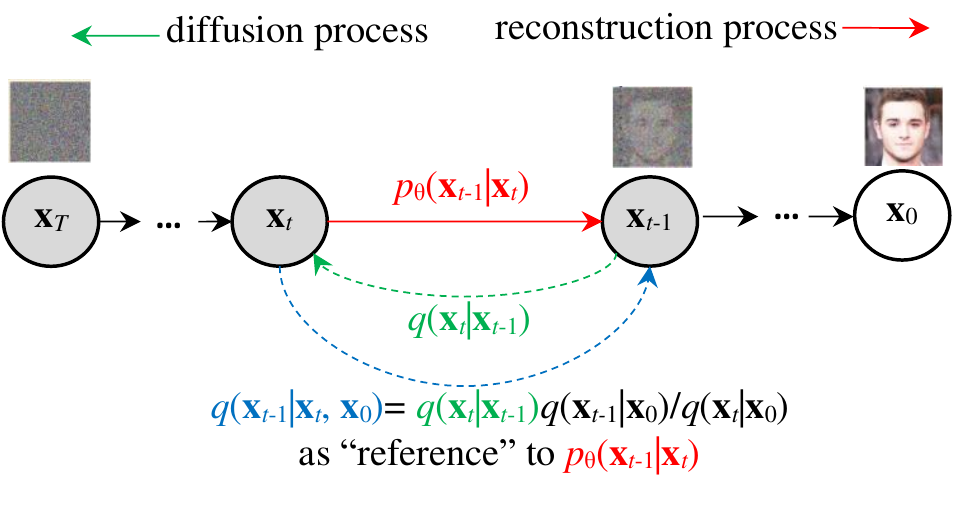}
  \caption{The Markov chain of forward diffusion (backward reconstruction) process of generating a sample by step-by-step adding (removing) noise. Image adapted from \cite{DDPM_DBLP:journals/corr/abs-2006-11239}.}
  \label{fig:ddpm_two_processes}
\end{figure}

We briefly introduce the diffusion and reconstruction processes in \textbf{D}enoising \textbf{D}iffusion \textbf{P}robabilistic \textbf{M}odels (DDPM).
Given a data point $\textbf{x}_0$ sampled from a real data distribution $q(\textbf{x})$ ($\textbf{x}_0 \sim q(\textbf{x})$), \newcite{DDPM_DBLP:journals/corr/abs-2006-11239} define a \emph{forward diffusion process} in which small amount of Gaussian noise is added to sample $\textbf{x}_0$ in $T$ steps to obtain a sequence of noisy samples $\textbf{x}_0, ..., \textbf{x}_T$. A predefined (hyper-parameter) variance schedule $\{ \beta_t \in (0, 1) \}_{t=1}^T$ controls the step sizes:
\begin{align}
    q(\textbf{x}_t | \textbf{x}_{t-1}) & = \mathcal{N}(\textbf{x}_t; \sqrt{1-\beta_t}\textbf{x}_{t-1}, \beta_t \textbf{I}); \\
    q(\textbf{x}_{1:T} | \textbf{x}_0) & := \prod_{t=1}^Tq(\textbf{x}_t | \textbf{x}_{t-1}). \label{eq:q_distribution}
\end{align}
When $T \rightarrow \infty$, $\textbf{x}_T$ is equivalent to following an isotropic Gaussian distribution. Note that, there are no trainable parameters used in this forward diffusion process.

Let $\alpha_t = 1 - \beta_t$ and $\bar{\alpha}_t = \prod_{i=1}^t \alpha_i$, we can express an arbitrary step $t$'s diffused sample $\textbf{x}_t$ by the initial data sample $\textbf{x}_0$:
\begin{equation}
    \textbf{x}_t = \sqrt{\bar{\alpha}_t} \textbf{x}_0 + \sqrt{1 - \bar{\alpha}_t} \bm{\epsilon}_t. \label{eq:xt_x0_relation}
\end{equation}
Here, noise $\bm{\epsilon}_t \sim \mathcal{N}(0, \textbf{I})$ shares the same shape with $\textbf{x}_0$ and $\textbf{x}_t$.

In order to reconstruct from a Gaussian noise input $\textbf{x}_T \sim \mathcal{N}(0, \textbf{I})$, we need to learn a model $p_\theta$ to approximate the conditional probabilities to run the \emph{reverse diffusion (reconstruction) process}:
\begin{align}
    p_\theta(\textbf{x}_{t-1} | \textbf{x}_{t}) & = \mathcal{N}(\textbf{x}_{t-1}; \bm{\mu}_\theta(\textbf{x}_t, t), \bm{\Sigma}_\theta(\textbf{x}_t, t)); \nonumber \\
    p_\theta(\textbf{x}_{0:T}) & := p(\textbf{x}_T)\prod_{t=1}^Tp_\theta(\textbf{x}_{t-1} | \textbf{x}_{t}). \label{eq:p_theta_distribution}
\end{align}

Note that the reverse conditional probability is tractable by first applying Bayes' rule to three Gaussian distributions and then completing the ``quadratic component'' in the $\text{exp}(\cdot)$ function:
\begin{align}
    q(\textbf{x}_{t-1} | \textbf{x}_{t}, \textbf{x}_0) & = \mathcal{N}(\textbf{x}_{t-1}; \tilde{\bm{\mu}}_t(\textbf{x}_t, \textbf{x}_0), \tilde{\beta}_t\textbf{I}) \\
    & = q(\textbf{x}_{t} | \textbf{x}_{t-1}, \textbf{x}_0)\frac{q(\textbf{x}_{t-1} | \textbf{x}_0)}{q(\textbf{x}_{t} | \textbf{x}_0)} \\
    & \propto \text{exp}(-\frac{1}{2\tilde{\beta}_t}(\textbf{x}_{t-1} - \tilde{\bm \mu}_t)^2).
\end{align}
Here, variance $\tilde{\beta}_t$ is a scalar and mean $\tilde{\bm \mu}_t$ depends on $\textbf{x}_t$ and noise $\bm{\epsilon}_t$:
\begin{align}
    \tilde{\beta}_t & = \frac{1-\bar{\alpha}_{t-1}}{1-\bar{\alpha}_{t}}\beta_t; \\
    \tilde{\bm \mu}_t & = \frac{1}{\sqrt{\alpha_t}}(\textbf{x}_t - \frac{1-{\alpha}_{t}}{\sqrt{1-\bar{\alpha}_{t}}}\bm{\epsilon}_t).
\end{align}
Intuitively, $q(\textbf{x}_{t-1} | \textbf{x}_{t}, \textbf{x}_0)$ acts as a \emph{reference} to learn $p_\theta(\textbf{x}_{t-1} | \textbf{x}_{t})$. We can use the variational lower bound (VLB) to optimize the negative log-likelihood:
\begin{multline}
    -\text{log}p_\theta(\textbf{x}_0) \leq -\text{log}p_{\theta}(\textbf{x}_0) + \\ D_{\text{KL}}(q(\textbf{x}_{1:T}|\textbf{x}_0) \parallel p_\theta(\textbf{x}_{1:T}|\textbf{x}_0)).
\end{multline}

Using the definitions of $q(\textbf{x}_{1:T}|\textbf{x}_0)$ in Equation \ref{eq:q_distribution} and $p_\theta(\textbf{x}_{0:T})$ in Equation \ref{eq:p_theta_distribution}, a loss item $L_t$ ($1 \leq t \leq T-1$) is expressed by:
\begin{align}
    \mathcal{L}_t & = D_{\text{KL}}(q(\textbf{x}_{t}|\textbf{x}_{t+1}, \textbf{x}_0) \parallel p_\theta(\textbf{x}_{t}|\textbf{x}_{t+1})) \\
    & = \mathbb{E}_{\textbf{x}_0, \bm{\epsilon}_t} \left [ \frac{1}{2\parallel \bm{\Sigma}_\theta(\textbf{x}_t, t)\parallel_2^2} \parallel \tilde{\bm{\mu}}_t - \bm{\mu}_\theta(\textbf{x}_t, t)\parallel^2 \right ]. \nonumber
\end{align}
We further reparameterize the Gaussian noise term instead to predict $\bm{\epsilon}_t$ from time step $t$'s input $\textbf{x}_t$ and use a simplified objective that ignores the weighting term:
\begin{align}
    \mathcal{L}_t^{\text{simple}} & = \mathbb{E}_{t \sim [1, T], \textbf{x}_0, \bm{\epsilon}_t} \left [ \parallel \bm{\epsilon}_t - \bm{\epsilon}_\theta(\textbf{x}_t, t)  \parallel^2 \right ] 
\end{align}

\section{Reversible Duplex Conformer}\label{sec:rdc_reversible_duplex_conformer}

In this paper, we extend the widely used Conformer \cite{conformer_gulati20_interspeech} architecture for encoding the speech signals into dense and compact representations of both ends. Conformer has achieved impressive results in supervised ASR by leveraging transformer's capturing of content-based \emph{global} interactions and convolutional neural network's exploiting of \emph{local} features. In Conformer, two macaron-like FFN layers with half-step residual connections sandwich the MHSA and convolution (CNN) modules followed by a post layer normalization. Besides supervised ASR, Conformer has also been successfully used in self-supervised Wav2Vec \cite{wav2vec_schneider19_interspeech, wav2vec2_NEURIPS2020_92d1e1eb} pretraining for downstream application tasks' fine-tuning.

\begin{figure}[t]
  \centering
  \includegraphics[width=7.5cm]{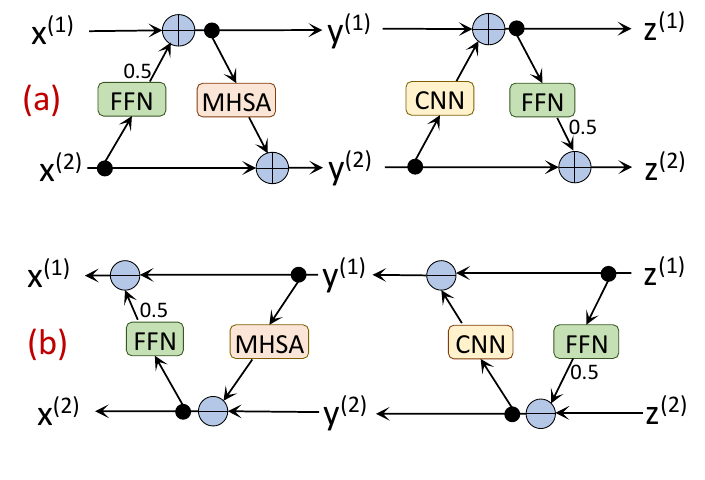}
  \caption{Forward (a) and reverse (b) building blocks for one layer in our reversible duplex Conformer. }
  \label{fig:reversible_duplex_conformer}
\end{figure}

\subsection{Forward and Reverse Building Blocks}

Following \cite{reverse_NIPS2017_f9be311e,reder_NEURIPS2021_afecc60f}, we split the $l$-th layer's (left-end) input tensor into two parts, $\mathbf{H}_{l-1}=[\mathbf{x}^{(1)}; \mathbf{x}^{(2)}]$. The (right-end) output tensor is split in the same way, $\mathbf{H}_{l}=[\mathbf{z}^{(1)}; \mathbf{z}^{(2)}]$. Thus, the forward target of this REDER-style Conformer layer is $[\mathbf{x}^{(1)}; \mathbf{x}^{(2)}] \mapsto [\mathbf{z}^{(1)}; \mathbf{z}^{(2)}]$. 

We introduce two intermediate tensors, $\mathbf{y}^{(1)}$ and $\mathbf{y}^{(2)}$, for intuitive understanding and mathematical convenient. Both Conformer's four sub-modules (two FFNs, one MHSA and one CNN) and four residual connections are kept in our reversible duplex Conformer. 

Figure \ref{fig:reversible_duplex_conformer} depicts the forward (a) and reverse (b) building blocks for one layer in our proposed reversible duplex Conformer. 
In Figure \ref{fig:reversible_duplex_conformer}, the reverse block is a mirror of the forward block with symmetrical network connections and subtract residual connections. The forward block can be formally expressed as follows:
\begin{align}
    \mathbf{y}^{(1)} & = \mathbf{x}^{(1)} + 0.5 \times \text{FFN}(\mathbf{x}^{(2)}); \\
    \mathbf{y}^{(2)} & = \mathbf{x}^{(2)} + \text{MHSA}(\mathbf{y}^{(1)}); \\
    \mathbf{z}^{(1)} & = \mathbf{y}^{(1)} + \text{CNN}(\mathbf{y}^{(2)}); \\
    \mathbf{z}^{(2)} & = \mathbf{y}^{(2)} + 0.5 \times \text{FFN}(\mathbf{z}^{(1)}).
\end{align}
Symmetrically, the reverse block is expressed by:
\begin{align}
    \mathbf{y}^{(2)} & = \mathbf{z}^{(2)} - 0.5 \times \text{FFN}(\mathbf{z}^{(1)}); \\
    \mathbf{y}^{(1)} & = \mathbf{z}^{(1)} - \text{CNN}(\mathbf{y}^{(2)}); \\
    \mathbf{x}^{(2)} & = \mathbf{y}^{(2)} - \text{MHSA}(\mathbf{y}^{(1)}); \\
    \mathbf{x}^{(1)} & = \mathbf{y}^{(1)} - 0.5 \times \text{FFN}(\mathbf{x}^{(2)}).
\end{align}

We employ Layer Normalization (LN) \cite{ln_Ba2016LayerN} at the beginning of each module, i.e., PreLN \cite{preln_10.5555/3524938.3525913}. The FFN module processes the input tensor $\mathbf{x}$ by six components:
\begin{equation}
    \text{FFN}(\mathbf{x}) = p_2 \circ \bm{W}_2 \circ p_1 \circ \text{SiLU} \circ \bm{W}_1 \circ \text{LN}(\mathbf{x}). \nonumber
\end{equation}
Here, $\circ$ means a layer takes $\circ$'s right-hand-side network's output (e.g., $\text{LN}(\mathbf{x})$) as the input of $\circ$'s left-hand-side network (e.g., $\bm{W}_1$ to perform $\bm{W}_1(\text{LN}(\mathbf{x}))$). $\bm{W}_1$ and $\bm{W}_2$ are two linear layers that preforms $h \mapsto 4h$ and $4h \mapsto h$ linear projections, respectively. Two dropout layers $p_1$ and $p_2$ are used. The Sigmoid Linear Unit (SiLU) \cite{silu_DBLP:journals/corr/ElfwingUD17} activation function is inserted between the two linear layers.
The MHSA module contains three components:
\begin{equation}
    \text{MHSA} = p \circ \text{Attention} \circ \text{LN}(\mathbf{x}). \nonumber
\end{equation}
We use multi-head attention with relative positional embedding \cite{relative_shaw2018relpos} for the ``Attention'' component. Note that, the attention module is extendable to cross-attention cases where a source sequence's encoded representation acts as memory (i.e., key and value) to the target sequence. Finally, the CNN module utilizes two types of convolutions, pointwise (PW) and 1D depthwise (DW), to capture local-range dependencies of the input speech. The idea of employing attention for global context modeling and convolution for local context modeling is also inspired by the long-short range attention mechanism used in the lite transformer \cite{wu2020lite}. Formally, 
\begin{align}
    \text{CNN}(\mathbf{x}) = p & \circ \text{PW}_2 \circ \text{Swish} \circ \text{BN} \nonumber
 \\ & \circ \text{DW} \circ \text{Glu} \circ \text{PW}_1 \circ \text{LN}(\mathbf{x}). \nonumber
\end{align}
Here, BN stands for batch normalization. Two types of activation functions, Glu \cite{glu_DBLP:journals/corr/DauphinFAG16} and Swish \cite{swish_DBLP:journals/corr/abs-1710-05941}, are inserted between convolution networks.

\subsection{Symmetric Network Architecture}

\begin{figure}[t]
  \centering
  \includegraphics[width=7.5cm]{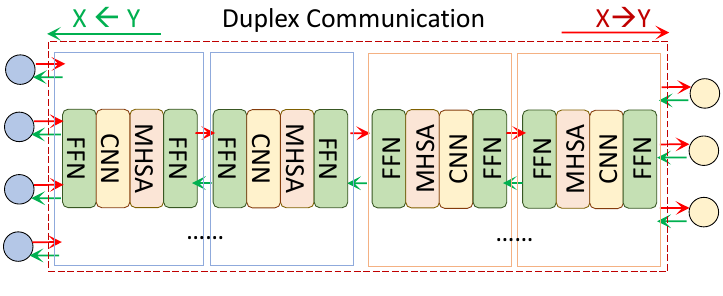}
  \caption{Symmetric architecture using reversible duplex Conformer building blocks for duplex speech-to-speech translation. }
  \label{fig:reversible_duplex_conformer_speech2speech}
\end{figure}

As depicted in Figure \ref{fig:reversible_duplex_conformer_speech2speech}, the forward and reverse building blocks are arranged symmetrically in the whole architecture to achieve homogeneous computations. Specifically, in the $L$ building blocks, the 1-st to $L/2$-th layers are set to be reverse blocks whereas the $(L/2+1)$-th to $L$-th layers be the regular forward form:
\begin{align}
    f_\theta^\rightarrow(\mathbf{x}) = \mathcal{F}_L & \circ \cdots \circ \mathcal{F}_{L/2+1} \nonumber \\
    & \circ \mathcal{F}_{L/2}^{-1} \circ \cdots \circ \mathcal{F}_1^{-1}(\mathbf{x}); \nonumber \\
    f_\theta^\leftarrow(\mathbf{z}) = \mathcal{F}_1 & \circ \cdots \circ \mathcal{F}_{L/2} \nonumber \\
    & \circ \mathcal{F}_{L/2+1}^{-1} \circ \cdots \circ \mathcal{F}_L^{-1}(\mathbf{z}). \nonumber
\end{align}
This design makes our reversible duplex Conformer to be homogeneous: the forward computational operation chain reads as a palindrome string $<fcmf\cdots fcmf | fmcf\cdots fmcf>$ and so does the reverse chain, where $f, m, c$ denotes FFN, MHSA and CNN, respectively. 

There are several selections of input types of the source and target ends in Figure \ref{fig:reversible_duplex_conformer_speech2speech}. \newcite{popuri2022enhanced} explores self-supervised pretrained models such as (1) wav2vec2 \cite{wav2vec2_NEURIPS2020_92d1e1eb} to encode the source speech and (2) Unit mBART \cite{mBART_DBLP:journals/corr/abs-2001-08210} to encode the target discrete units \cite{s2s_unit_DBLP:journals/corr/abs-2107-05604}, and then translate source speech into target clustered units through fine-tuning. The generated discrete unit sequence is then sent to an independently trained ``text''-to-speech (TTS) model to obtain the final waves. 

In this paper, we follow the usage of discrete units that are generated by first using pretrained HuBERT \cite{hubert_DBLP:journals/corr/abs-2106-07447} to encode the target speech and then perform k-means clustering. Then, we use the DiffWave \cite{diffwave_Kong2020DiffWaveAV} vocoder to generate the final waves.

\section{Duplex Diffusion Model}\label{sec:ddm_duplex_diffusion_model}

\begin{algorithm}[t!]  
    \caption{Duplex Diffusion Model (DDM) Training Algorithm - One Step}  
    \label{alg:duplex_diffusion_model_s2s}
    Given: $\mathbf{x}$, $\mathbf{y}$, $\mathcal{E}_{\mathbf{x}}$, $\mathcal{E}_{\mathbf{y}}$\\
    $\mathbf{x}_0 = \mathcal{E}_{\mathbf{x}}(\mathbf{x})$ $\triangleright$ encode by pretrained wav2vec models \\
    $\mathbf{y}_0 = \mathcal{E}_{\mathbf{y}}(\mathbf{y})$ $\triangleright$ encode by pretrained wav2vec models \\
    $t \sim \text{Uniform}({1, ..., T})$ \\
    $\mathbf{\epsilon}_{\mathbf{x}}\sim \mathcal{N}_{\mathbf{x}}(\mathbf{0}, \mathbf{I}), \mathbf{\epsilon}_{\mathbf{y}} \sim \mathcal{N}_{\mathbf{y}}(\mathbf{0}, \mathbf{I})$ \\
    $\mathbf{x}_t = \sqrt{\bar{\alpha}_{t, \mathbf{x}}} \mathbf{x}_0 + \sqrt{1 - \bar{\alpha}_{t,\mathbf{x}}} \mathbf{\epsilon}_{\mathbf{x}}$ \\
    $\mathbf{y}_t = \sqrt{\bar{\alpha}_{t, \mathbf{y}}} \mathbf{y}_0 + \sqrt{1 - \bar{\alpha}_{t,\mathbf{y}}} \mathbf{\epsilon}_{\mathbf{y}}$ \\

    $\mathbf{\epsilon}^\theta_{\mathbf{x}} = \overleftarrow{\mathcal{M}_\theta}(\mathbf{x}_t, t, \mathbf{y}_0)$ $\triangleright$ reverse, given $\mathbf{y}_0$ \\
    $\mathbf{\epsilon}^\theta_{\mathbf{y}} = \overrightarrow{\mathcal{M}_\theta}(\mathbf{y}_t, t, \mathbf{x}_0)$ $\triangleright$ forward, given $\mathbf{x}_0$ \\
    $\mathcal{L}_{\text{DDM}} = \lambda_1 \parallel \mathbf{\epsilon}_{\mathbf{x}} - \mathbf{\epsilon}^\theta_{\mathbf{x}} \parallel ^2 + \lambda_2 \parallel \mathbf{\epsilon}_{\mathbf{y}} - \mathbf{\epsilon}^\theta_{\mathbf{y}} \parallel ^2 $ \\
\end{algorithm}

Cycle consistency has been utilized in textual neural machine translation \cite{reder_NEURIPS2021_afecc60f} and image-to-image translation \cite{ddib_Su2022DualDI}. In this paper, we propose a duplex diffusion model that alternatively optimizes both directions by two diffusion processes.

The training algorithm is described in Algorithm \ref{alg:duplex_diffusion_model_s2s}. Generally, we borrow DDPM \cite{DDPM_DBLP:journals/corr/abs-2006-11239}'s architecture and extend it to a duplex scenario where sequences of two ends are diffused alternatively during training. At the beginning, the source sequence $\mathbf{x}$ and target sequence $\mathbf{y}$ are encoded into dense representations by pretrained wav2vec models $\mathcal{E}_{\mathbf{x}}, \mathcal{E}_{\mathbf{y}}$ through self-supervised learning on monolingual datasets, respectively. Then, time $t$ and two normal Gaussian noise signals $\mathbf{\epsilon}_{\mathbf{x}}, \mathbf{\epsilon}_{\mathbf{y}}$ are sampled. Note that the lengths of the source and target sequences are diverse. We pre-define two variance schedules $\{ \beta_{t, \mathbf{x}} \in (0, 1) \}_{t=1}^T$ and $\{ \beta_{t, \mathbf{y}} \in (0, 1) \}_{t=1}^T$, for the source and target languages, respectively. Thus, we have $\alpha_{t, \mathbf{x}} = 1 - \beta_{t, \mathbf{x}}$, $\bar{\alpha}_{t, \mathbf{x}} = \prod_{i=1}^t \alpha_{i, \mathbf{x}}$, $\alpha_{t, \mathbf{y}} = 1 - \beta_{t, \mathbf{y}}$ and $\bar{\alpha}_{t, \mathbf{y}} = \prod_{i=1}^t \alpha_{i, \mathbf{x}}$, as used in Algorithm \ref{alg:duplex_diffusion_model_s2s}.

The variance schedules, initial sequence representations and normal Gaussian noises work together to give us diffused representations, $\mathbf{x}_t$ and $\mathbf{y}_t$, respectively. They are then sent to the reversible duplex Conformer architecture $\mathcal{M_\theta}$ (Figure \ref{fig:reversible_duplex_conformer_speech2speech}) to predict the noises. 

Originally in Figure \ref{fig:reversible_duplex_conformer_speech2speech}, we are intended to produce $\mathbf{x}_0$ from $\mathbf{y}_0$ in the reverse process of $\overleftarrow{\mathcal{M}_\theta}$. Now, we have two additional inputs, $t$ and $\mathbf{x}_t$. The output also changes from predicting $\mathbf{x}_0$ to estimating $\mathbf{\epsilon}_\mathbf{x}^\theta$ which shares the same shape with $\mathbf{x}_0$. 

We thus have two ways to organize the network $\overleftarrow{\mathcal{M}_\theta}$: (1) reuse Figure \ref{fig:reversible_duplex_conformer_speech2speech}'s architecture and predicting $\mathbf{\epsilon}_\mathbf{x}^\theta$ from $\mathbf{y}_0$ by taking $\mathbf{x}_t$ as the ``memory'' which acts as key and value in the cross-attention network in Conformer, or (2) follow traditional stable diffusion models \cite{stablediffusion_DBLP:journals/corr/abs-2112-10752} and predict $\mathbf{\epsilon}_\mathbf{x}^\theta$ from $\mathbf{x}_t$ by taking $\mathbf{y}_0$ as the conditional ``memory''. That is, in the MHSA function, we set query $\mathbf{x}_t$ to be and key/value to be $\mathbf{y}_0$, $\text{MHSA}(q=\mathbf{x}_t, k=\mathbf{y}_0, v=\mathbf{y}_0)$, so that the identical lengths of $q=\mathbf{x}_t$ and $\mathbf{\epsilon}_\mathbf{x}^\theta$ are ensured. Note that, in the second choice used in our experiments, we are not limited to use a reversible duplex Conformer, i.e., any transformer architecture with cross-attention are applicable. These two options still hold during inferencing from given $\mathbf{y}_0$, $T$, and $\mathbf{x}_T$ to iteratively reconstruct $\mathbf{x}_0$.

Since the lengths of the source and target sequences are diverse, we follow textual duplex translation \cite{reder_NEURIPS2021_afecc60f} and double the source end's length by a upsampling convolutional network. 

We only describe the reverse process $\overleftarrow{\mathcal{M}_\theta}$ and the forward process $\overrightarrow{\mathcal{M}_\theta}$ shares the similar strategies. To achieve a full cycle consistency, predicting the target Gaussian noise from source sequence by taking target noisy sequence as the ``conditional memory'' is more appropriate in current scenario setting so that both translation directions are achieved in one duplex diffusion model.

After the reverse and forward processes, we can compute the MSE losses of between the two pairs of reference and predicted noises. They are interpolated together by hyper-parameter weights $\lambda_1$ (=0.5) and $\lambda_2$ (=0.5) to the final loss $\mathcal{L}_{\text{DDM}}$ to be optimized. 

\section{Training}\label{sec:training_losses_and_details}

Our reversible duplex Conformer is largely inspired by REDER \cite{reder_NEURIPS2021_afecc60f}. The novel parts are that (1) we select and reconstruct convolution-enhanced Conformer \cite{conformer_gulati20_interspeech} to synthetically capture global information by attentions and local context by convolutions, and (2) we extend from textual duplex machine translation to (dense) duplex speech-to-speech translation. When training our reversible duplex Conformer, we borrow and adapt the losses that are used in REDER to fit our scenario. 

In REDER, three types of losses were used. The first loss is to model the variable-length of source and target sequences by a latent alignment approach, i.e., the Connectionist Temporal Classification (CTC) \cite{ctcloss_10.1145/1143844.1143891}. Starting from the conditional independence assumption, CTC is capable of efficiently (by dynamic programming) finding all valid (yet monotonic) alignments $\mathbf{a}$ which derives from the target $\mathbf{y}$ by allowing consecutive repetitions and inserting blank tokens. The CTC loss is defined by:
\begin{equation}
    \mathcal{L}_{\text{CTC}} = - \text{log}p_{\text{CTC}}(\mathbf{y}|\mathbf{x}; \theta) = -\text{log}\sum_{\mathbf{a}}p_\theta(\mathbf{a}|\mathbf{x}). \nonumber
\end{equation}
We adapt this loss for speech translation when the target are clustered unit sequences. We use the MSE loss instead when the target is a sequence of mel-spectrogram. Also, to ensure the source sequence is always longer than the target sequence, we upsample the source sequences by convolutional layers before sending them to the reversible duplex Conformer.

The second loss measures the layer-wise forward-backward agreement (fba, measured by cosine similarity) of between the forward $l$-th layer's representation $\overrightarrow{\mathbf{H}}_l$ = $\mathcal{F}_l(\overrightarrow{\mathbf{H}}_{l-1})$ and the reverse representation $\overleftarrow{\mathbf{H}}_l$ = $\mathcal{F}_l(\overleftarrow{\mathbf{H}}_{l+1})$. Thus, 
\begin{equation}
    \mathcal{L}_{\text{fba}}(\mathbf{y}|\mathbf{x}; \theta) = \frac{1}{L}\sum_{l=1}^L\left \{1-\text{cos}(\overrightarrow{\mathbf{H}}_l, \texttt{sg}(\overleftarrow{\mathbf{H}}_l))\right \}, \nonumber
\end{equation}
where $\texttt{sg}$ denotes the stop-gradient operation.

The third loss explicitly describe the cycle consistency of a pair of seq2seq tasks, i.e., we minimize the distance between the original $\mathbf{x}$ and its reconstruction ${f_\theta^\leftarrow}({f_\theta^\rightarrow}(\mathbf{x}))$ by,
\begin{equation}
    \mathcal{L}_{\text{cc}}(\mathbf{x}; \theta) = \texttt{distance}(\mathbf{x}, {f_\theta^\leftarrow}({f_\theta^\rightarrow}(\mathbf{x}))). \nonumber
\end{equation}
For speech translation, the source sequence can be expressed by mel-spectrogram or clustered units so that MSE loss or CTC loss can be applied to them, respectively. Finally, these three types of losses are doubled to two directions and interpolated together for the final loss. That is, when predicting discrete units, the final loss function is:
\begin{align}
    \mathcal{L}_{\text{unit}} = &\ w_1*\mathcal{L}_{\text{CTC}}(\mathbf{y}|\mathbf{x}) + w_2*\mathcal{L}_{\text{CTC}}(\mathbf{x}|\mathbf{y}) \nonumber \\
    & + w_3*\mathcal{L}_{\text{fba}}(\mathbf{y}|\mathbf{x}) + w_4*\mathcal{L}_{\text{fba}}(\mathbf{x}|\mathbf{y}) \nonumber \\
    & + w_5*\mathcal{L}_{\text{cc}}(\mathbf{y}) + w_6*\mathcal{L}_{\text{cc}}(\mathbf{x}). \nonumber
\end{align}
When predicting mel-spectrograms, the final loss function is:
\begin{align}
    \mathcal{L}_{\text{mel}} = &\ w_1*\mathcal{L}_{\text{MSE}}(\mathbf{y}|\mathbf{x}) + w_2*\mathcal{L}_{\text{MSE}}(\mathbf{x}|\mathbf{y}) \nonumber \\
    & + w_3*\mathcal{L}_{\text{fba}}(\mathbf{y}|\mathbf{x}) + w_4*\mathcal{L}_{\text{fba}}(\mathbf{x}|\mathbf{y}) \nonumber \\
    & + w_5*\mathcal{L}_{\text{cc}}(\mathbf{y}) + w_6*\mathcal{L}_{\text{cc}}(\mathbf{x}). \nonumber
\end{align}
We reuse the default hyper-parameter values described in REDER \cite{reder_NEURIPS2021_afecc60f} for setting weights $w_1$ to $w_6$.

In our experiments, we first train the reversible duplex Conformer architecture by a predefined $K_1$ (=200,000) iterations and then apply the duplex diffusion training algorithm as shown in Algorithm \ref{alg:duplex_diffusion_model_s2s}. After another predefined $K_2$ (=200,000) iterations, we fix the diffusion processes and focus on updating the reversible duplex Conformer only so that traditional search algorithms such as beam search can be used for seeking target hypotheses. 

\section{Experimental Setups}

\subsection{Data}

To compare with state-of-the-art baselines' reported results, we align with UnitY \cite{unity_s2s_2022} and use three S2ST datasets: (1) Fisher Es$\rightarrow$En \cite{post-etal-2013-improved} with 170-hour Spanish (Es) conversational telephone speech with textual transcriptions in Es and En. The English speech is synthesized by a single-female-speaker TTS model. (2) CVSS-C \cite{jia-etal-2022-cvss}, a public multilingual S2ST corpus from CoVoST2 \cite{DBLP:journals/corr/abs-2007-10310}. Again, a single-female-speaker TTS model is employed to synthesize the target speech. (3) Multi-domain En$\leftrightarrow$Es corpora \cite{popuri2022enhanced}. We follow \cite{unity_s2s_2022} to collect 1983-hour source speech for En$\rightarrow$Es and 1404-hour source speech for Es$\rightarrow$En. 

\subsection{Pre-training and Pre-processing}

We use the pretrained wav2vec2.0 \cite{wav2vec2_NEURIPS2020_92d1e1eb} with a 24-layer Conformer \cite{conformer_gulati20_interspeech} self-trained on the Libri-Light dataset \cite{libri_light_DBLP:journals/corr/abs-1912-07875}, HuBERT \cite{hubert_DBLP:journals/corr/abs-2106-07447}, mHuBERT \cite{popuri2022enhanced}, and mBART \cite{liu-etal-2020-multilingual-denoising} given in Table 9 of \cite{unity_s2s_2022}.

For acoustic feature extraction, discrete unit extraction (100 clusters) and text normalization (e.g., for evaluation score computing), we follow \cite{popuri2022enhanced, unity_s2s_2022}.

\subsection{Vocoder}

Instead of using the HiFi-GAN vocoder \cite{Kong2020HiFiGANGA, DBLP:journals/corr/abs-2104-00355} which converts mel-spectrograms or discrete units to waveform sequences for TTS and direct speech-to-spectrogram/unit models, we borrow a comparable diffusion based vocoder, DiffWave \cite{diffwave_Kong2020DiffWaveAV}, for reconstructing waveforms from spectrogram or unit sequences.

\subsection{Training and Decoding Configurations}

We implement our models based on the Fairseq toolkit\footnote{\url{https://github.com/facebookresearch/fairseq}} \cite{ott-etal-2019-fairseq}. All our models are optimized with a mixed precision training for footprint saving. Our reversible duplex Conformer uses the settings of Conformer-Large with 135.1M parameters \cite{conformer_gulati20_interspeech}. The two diffusion variance schedules used in our duplex diffusion model follow stable diffusion \cite{stablediffusion_DBLP:journals/corr/abs-2112-10752}. We use a NVIDIA DGX-A100*8 workstation to perform the training with a total of 2,500 GPU hours.

During inferencing, we set the beam size to be 10 which aligns with most of the baselines for fair comparison. Other configurations not mentioned here follow their default settings in their open-source repositories.

\subsection{Evaluation}

We use a pre-trained ASR model to transcribe the target speech into texts and then calculate 4-gram BLEU scores \cite{papineni-etal-2002-bleu}, denoted as ASR-BLEU. The target languages' ASR models are fine-tuned from pretrained wav2vec2.0 \cite{wav2vec2_NEURIPS2020_92d1e1eb} models with the CTC objective \cite{ctcloss_10.1145/1143844.1143891} when we taking discrete unit sequences as the prediction target. The same criterion has been used in \cite{unity_s2s_2022}. 

\section{Experimental Results}

\subsection{Fisher Es$\rightarrow$ En}

\begin{table}
  \centering
  \begin{tabular}{lccc}
  \hline
  \textbf{Model} & \textbf{dev} & \textbf{dev2} & \textbf{test} \\
  \hline
  ASR-MT-TTS & 42.1 & 43.5 & 43.9 \\
  S2TT-TTS, C & 47.8 & 48.9 & 48.3 \\
  S2TT-TTS, C-w2v2 & 51.0 & 52.2 & 52.1 \\
  \hline
  S2Sp-Tn, C & 43.9 & 44.4 & 43.8 \\
  S2Sp-Tn, C-w2v2 & 45.5 & 47.6 & 46.3 \\
  S2Sp-Tn2+, C & 50.4 & 51.1 & 50.8 \\
  S2Sp-Tn2+, C-w2v2 & 58.4 & 59.5 & 58.6 \\
  \hdashline
  S2Sp-RDC (Ours) & 46.1 & 47.3 & 47.0 \\
  S2Sp-RDC, w2v2 & 50.7 & 51.5 & 51.0\\
  S2Sp-DDM (Ours) & 52.4 & 55.1 & 54.8\\
  S2Sp-DDM, w2v2 & \textbf{58.9} & \textbf{59.8} & \textbf{59.1}\\
  \hline
  S2U, C & 46.2 & 47.6 & 47.4 \\
  S2U, C-w2v2 & 53.4 & 53.9  & 53.7 \\
  UnitY, C & 50.5& 51.6& 51.4 \\
  UnitY, C-w2v2 & 55.1& 56.5 &55.9 \\
  \hdashline
  S2U-RDC (Ours) & 48.1 & 49.0 & 48.5\\
  S2U-RDC, w2v2 & 50.8 & 52.1 & 51.8\\
  S2U-DDM (Ours) & 52.2 & 53.6 & 53.1\\
  S2U-DDM, w2v2 & \textbf{56.3} & \textbf{58.0} & \textbf{57.4} \\
  \hline
  \end{tabular}
  \caption{ASR-BLEU (\%) on the Fisher Es$\rightarrow$En corpus. S2Sp = speech-to-spectrogram, S2U = speech-to-unit, Tn = Translatotron, C = Conformer, RDC = reversible duplex Conformer (Section \ref{sec:rdc_reversible_duplex_conformer}), DDM = duplex diffusion model (Section \ref{sec:ddm_duplex_diffusion_model}), and w2v2 = wav2vec2.0.}
  \label{tab:fisher_es2en}
\end{table}

In Table \ref{tab:fisher_es2en}, we compare the ASR-BLEU scores of our systems (RDC and DDM) with three cascaded systems, four speech-to-spectrogram baselines which are variants of Translatotron \cite{Translatotron_Jia2019DirectST, Jia2021Translatotron2H}, and four speech-to-unit baselines which are variants of \cite{s2s_unit_DBLP:journals/corr/abs-2107-05604} and UnitY \cite{unity_s2s_2022}. Baseline results are originally listed in \cite{unity_s2s_2022}. 

We use RDC to denote our reversible duplex Conformer architecture that are trained in a similar way with textual REDER \cite{reder_NEURIPS2021_afecc60f}. Our DDM further ``boost'' the quality of pretrained RDC models by bidirectional diffusion processes and can be recognized as an integration of the diffusion framework with RDC. Of the three categories, S2Sp and S2U achieved significantly better ($p<0.01$) ASR-BLEU scores than the three traditional cascaded systems. In the S2Sp paradigm, our ``S2Sp-DDM, w2v2'' model achieves comparable results with the best baseline ``S2Sp-Tn2+, C-w2v2''. In the S2U paradigm, our model ``S2U-DDM, w2v2'' achieves significantly better ($p<0.05$) results than the best baseline ``UnitY, C-w2v2'', with 1.2\%, 1.5\% and 1.5\% absolute ASR-BLEU points. These reflects that our proposed duplex seq2seq learning can be boosted by the bidirectional diffusion processes to better capture the translation distributions of among the source and target sides. In addition, wav2vec2.0 acts as an essential component for all the model variants.

Table \ref{tab:fisher_es2en} also lists four variants of our models for ablation study. When we only use S2U-RDC, it performs better than the S2U+Conformer baseline. However, this advantage disappears when w2v2 is further employed to these two variants. S2U-RDC also performs relatively worse than UnitY which employs two pass decoding of basing on texts and units whereas our S2U-RDC uses units only. These reflect that, (1) additional textual information brings better results than duplex training, (2) diffusion processes can partly ``hedge'' the benefits from two-pass decoding used in UnitY and enhance the performance of duplex translations.

\subsection{CVSS-C}

\begin{table}
  \centering
  \begin{tabular}{lcccc}
  \hline
  \textbf{Model} & \textbf{Avg.} & \textbf{High} & \textbf{Mid} & \textbf{Low} \\
  \hline
  S2TT-TTS, ASR & 12.7 & 30.7 & 18.3 & 4.4 \\
  S2TT-TTS, w2v-b & 13.2 & 21.3 & 16.1 & 9.2 \\
  \hline
  S2Sp-Tn2, w2v-b & 17.9 & 32.5 & 22.9 & 10.9 \\
  S2Sp-Tn2+, w2v-b &  20.8 & 31.6  & 25.4 & \textbf{15.4} \\
  \hdashline
  S2Sp-RDC (Ours) & 18.2 & 32.4 & 22.1 & 10.2 \\
  S2Sp-DDM (Ours) & \textbf{22.1} &\textbf{33.5} & \textbf{27.4} & 15.2 \\
  \hline
  S2U, w2v-b & 20.8 & 31.6 & 25.4 & 15.4 \\
  UnitY, w2v-b & 24.5 & 34.6 & 28.9 & 19.3  \\
  \hdashline
  S2U-RDC (Ours) & 22.1 & 32.5 & 27.1 & 17.8 \\
  S2U-DDM (Ours) & \textbf{24.9} & \textbf{35.2} & \textbf{30.2} & \textbf{20.4} \\
  \hline
  \end{tabular}
  \caption{ASR-BLEU (\%) on the CVSS-C corpus. ASR = ASR pretraining, w2v-b = wav2vec BERT.}
  \label{tab:cvssc_results}
\end{table}

The CVSS-C corpus's ASR-BLEU scores of six baselines from three categories and our models are listed in Table \ref{tab:cvssc_results}. We observe almost the same tendencies with the result comparisons in the Fisher task (Table \ref{tab:fisher_es2en}). The best baseline is still the two-pass UnitY model enhanced by a pretrained wav2vec-BERT model. Our S2U-DDM model improves UnitY by 0.4\% ASR-BLEU points on average, comparable yet not significant. 

\subsection{Multi-domain En$\leftrightarrow$Es}

\begin{table}
  \centering
  \begin{tabular}{lcc}
  \hline
  \textbf{Model En$\rightarrow$Es} & \textbf{Europarl-ST} & \textbf{MuST-C}   \\
  \hline
  ASR-MT-TTS & 36.8 & 30.8  \\
  S2TT-TTS & 36.4 & 33.4  \\
  \hline
  S2Sp-Tn2+ & 35.6 & 33.5  \\
  S2Sp-Tn2+, mB &  36.9 & \textbf{34.3}   \\
  \hdashline
  S2Sp-RDC (Ours) &  35.1 & 32.7 \\
  S2Sp-DDM (Ours) &  \textbf{37.2} & \textbf{34.3} \\
  \hline
  UnitY & 35.1 & 33.7 \\
  UnitY, mB & 35.3 & 34.1 \\
  \hdashline
  S2U-RDC (Ours) &  34.7 & 32.6\\
  S2U-DDM (Ours) &  \textbf{35.8} & \textbf{34.5} \\
  \hline
  \hline
  \textbf{Model Es$\rightarrow$En} & \textbf{CoVoST-2} & \textbf{Europarl-ST}   \\
  \hline
  ASR-MT-TTS & 32.9 & 34.2   \\
  S2TT-TTS & 37.2 & 34.0  \\
  \hline
  S2Sp-Tn2+ & 37.0 & 23.4  \\
  S2Sp-Tn2+, mB &  \textbf{37.2} & 23.7   \\
  \hdashline
  S2Sp-RDC (Ours) &  34.5 & {30.6}  \\
  S2Sp-DDM (Ours) &  37.1 & \textbf{32.8}   \\
  \hline
  UnitY & 35.4 & 30.8 \\
  UnitY, mB & 36.4 & 33.1 \\
  \hdashline
  S2U-RDC (Ours) &  35.1 & 31.2  \\
  S2U-DDM (Ours) & \textbf{36.7} & \textbf{34.0} \\
  \hline
  \end{tabular}
  \caption{ASR-BLEU (\%) on the multi-domain En$\leftrightarrow$Es tasks. mB = mBART.}
  \label{tab:multi_domain_en2es_es2en}
\end{table}

The bidirectional multi-domain En$\leftrightarrow$Es results are listed in Table \ref{tab:multi_domain_en2es_es2en}. We again compare with six state-of-the-art baselines in three categories. On both directions, our model variants meet the best performances on the two test sets. We notice that the baselines perform less stable under the Europarl-ST corpus with ASR-BLEU ranges from 23.4\% to 34.2\%. In the S2Sp scenario, both our RDC and DDM variants perform significantly better ($p<0.01$) than the two baselines. Our S2U-DDM variant performs significantly better ($p<0.05$) than UnitY and is comparable to the best cascaded system. Note that we only require one run training for bidirectional translations. 

\subsection{Inference Speed}

We use a NVIDIA DGX-A100*8 workstation to perform the inferencing comparison without additional engineering optimization. We randomly select 500 utterances from the multi-domain Es$\rightarrow$En dev set. For end-to-end S2ST inferencing, our final RDC with one-pass decoding achieved 1.72$\times$ decoding speed-ups over the best-performance UnitY \cite{unity_s2s_2022} baseline which requires a two-pass text+unit decoding.

\subsection{Human Evaluation}

Finally, we performed an audio-only human evaluation to evaluate the translation quality and acceptances of the best baseline UnitY and our DDM. For direct comparison, we use the mTEDx test with 989 samples. We obtained a mean translation quality score of 4.202(/5.0) which is comparable to UnitY's 4.197 and an acceptable ratio of 92.89\% which is also comparable to UnitY's 92.94\%.   

\section{Conclusion}

Aiming at effectively leveraging bidirectional supervision signals of speech-to-speech translation (S2ST), we have proposed two models for duplex S2ST, a reversible duplex Conformer and a duplex diffusion model. We compare with cascaded S2ST models, single/multi-pass speech-to-spectrogram/unit models and report significantly better or comparable ASR-BLEU and human-evaluated scores, with fewer training time and faster inference speed.  

\section{Limitations}\label{sec:limitations}

Our duplex diffusion model and reversible duplex Conformer architecture do not explicitly take re-ordering as an essential challenge. However, depicts the language pairs described in the experiments, there are languages such as English and Japanese which shares subject-verb-object (SVO) and subject-object-verb (SOV) word orders. These limit the scalability of our proposed methods and external pre-ordering \cite{zhao-etal-2018-exploiting,wu-etal-2011-extracting} or post-ordering \cite{10.1145/2518100} techniques on clustered units of speech should be taken into consideration in the future work.

Large-scale unlabeled speech data is required to train self-supervised wav2vec2.0 or HuBERT models. However, this is frequently not easy to collect. Moreover, it is even more difficult to collect paired speech-to-speech data and existing TTS models for generating speech from text are still under developing. These are pre-conditions of applying our proposed approaches.

Finally, we still need to train pair-by-pair for S2ST which is quadratic to the number of languages. Our approach is less effective than textual multilingual machine translation architecture in which linear number of translation models are required and achieved comparable results than pair-wise baselines. Multilingual S2ST requires novel training architectures and inferencing algorithms.

\section{Ethics Statement}

Our target is to build direct speech-to-speech translation systems with a duplex idea of training both directions in one run. We try our best to reuse existing pretrained wav2vec2.0, HuBERT, mHuBERT and mBART models to save energy consuming. In one run, we require much less GPU-hours for obtaining S2ST models for both direction usages. However, compared with textual duplex MT systems, pre-processing of speech signals still requires much higher costing of GPU-hours and as listed in our limitation section (Section \ref{sec:limitations}), smarter ways of multilingual S2ST architectures are preferred in the future to reduce the cost of energy from current quadratic to linear number of models. 

Generally, S2ST circumvents traditional cascaded systems which concatenate ASR, MT and TTS with high latency and high requirements of datasets. There are 3,000 around languages in the world who do not have their own writing systems or textual vocabularies. Through our duplex S2ST models, we hope to be friendly to these languages so that more and more languages can be covered.

\bibliography{anthology,custom}
\bibliographystyle{acl_natbib}


\end{document}